\documentclass[nohyperref]{article}

\usepackage{microtype}
\usepackage{graphicx}
\usepackage{subfigure}
\usepackage[svgnames]{xcolor}
\usepackage{booktabs} 
\usepackage{svg}

\usepackage{caption}
\usepackage{multirow}
\usepackage{hyperref}
\usepackage{xcolor}

\usepackage[accepted]{preprint}

\usepackage{amsmath}
\usepackage{amssymb}
\usepackage{mathtools}
\usepackage{amsthm}

\usepackage[capitalize,noabbrev]{cleveref}

\theoremstyle{plain}

\theoremstyle{definition}

\theoremstyle{remark}

\usepackage[textsize=tiny]{todonotes}

\icmltitlerunning{~ \hfill Object-Centric Scene Representations using Active Inference \hfill \thepage}

\begin{document}

\twocolumn[
\icmltitle{Object-Centric Scene Representations using Active Inference}



\icmlsetsymbol{equal}{*}

\begin{icmlauthorlist}
\icmlauthor{Toon Van de Maele}{ugent}
\icmlauthor{Tim Verbelen}{ugent}
\icmlauthor{Pietro Mazzaglia}{ugent}
\icmlauthor{Stefano Ferraro}{ugent}
\icmlauthor{Bart Dhoedt}{ugent}
\end{icmlauthorlist}

\icmlaffiliation{ugent}{IDLab, Ghent University, Belgium}

\icmlcorrespondingauthor{Toon Van de Maele}{toon.vandemaele@ugent.be}

\icmlkeywords{Scene Representation Learning, Active Inference, Active Vision}

\vskip 0.3in
]



\printAffiliationsAndNotice{}  

\begin{abstract}
Representing a scene and its constituent objects from raw sensory data is a core ability for enabling robots to interact with their environment. In this paper, we propose a novel approach for scene understanding, leveraging a hierarchical object-centric generative model that enables an agent to infer object category and pose in an allocentric reference frame using active inference, a neuro-inspired framework for action and perception. For evaluating the behavior of an active vision agent, we also propose a new benchmark where, given a target viewpoint of a particular object, the agent needs to find the best matching viewpoint given a workspace with randomly positioned objects in 3D. We demonstrate that our active inference agent is able to balance epistemic foraging and goal-driven behavior, and outperforms both supervised and reinforcement learning baselines by a large margin.
\end{abstract}

\section{Introduction}

Spatial scene understanding is a core ability for enabling robots to understand and interact with their environment and has been a long-standing challenge in computer vision. Humans naturally decompose scenes into object-centric representations and infer information about objects, their appearance, their constituent parts, as well as their pose and shape in the 3D space~\citep{Hinton79}. For example, when seeing a coffee cup, humans  immediately know where to reach for the handle, even when the handle is not directly in view.

In the past decade, advances in deep learning have enabled to devise systems that can distinguish objects in images, i.e. predicting segmentation masks~\citep{Minaee21}, or the object pose in 3D~\citep{xiang2017posecnn, du21} by supervised training on a dataset annotated with predefined object classes and poses. Also, unsupervised methods have been proposed to infer separate objects from one~\citep{Eslami2016, Bear2020} or multiple~\citep{chen_roots_2021} views. In order to have full 3D scene understanding, other approaches learn representations of complete 3D shapes by training on 3D CAD models of objects of interest~\citep{CVPR15_Wu}. Such representations can then be used to infer objects shape and pose in 3D from one or more RGB-D images~\citep{Sucar2020}. However, these methods typically operate on static inputs and do not allow the agent to interact with the scene.

In contrast, humans learn by actively engaging and interacting with the world~\citep{Karin2014}. A prevailing account of human perception is that the brain builds a generative model of the world, constantly explaining its observations~\citep{Friston2017}, i.e. active inference. In this regard, vision is cast as inverting a generative model of the scene, in order to infer its constituent objects, their appearance, and pose~\citep{parr_generative_2021}. This is consistent with findings in the brain, where visual inputs are processed by a dorsal (``where'') stream on the one hand, representing where an object is in the space, and a ventral (``what'') stream on the other hand, representing object identity~\citep{Mishkin1983}. 

In this paper, we propose a novel method for spatial scene understanding, using an agent that can actively move the camera in the scene. We endow the agent with an object-centric generative model, which is optimized by minimizing free energy. By also inferring actions that minimize expected free energy, the agent engages in active inference and balances goal-directed with explorative behavior. To evaluate our approach, we present a novel benchmark that casts active vision as a task in which an artifical agent needs to reach a goal observation. To demonstrate this, our paper proposes the following contributions:
\begin{itemize}
    \item We propose a novel hierarchical object-centric generative model that factorizes object identity from a location in both egocentric and allocentric reference frames (Section~\ref{sect:method}).
    \item We developed a new benchmark environment for evaluating the active vision problem. In this environment, the agent needs to find a particular object by actively engaging with the environment through moving the camera (Section~\ref{sect:env}).
     \item We compare performance both quantitatively and qualitatively against two other methods that adopt a supervised~\citep{xiang2017posecnn} and a reinforcement learning~\citep{lexa} approach respectively (Section~\ref{sect:results}) and show that we perform better on the active vision benchmark. 
    \item We demonstrate that driving active vision through the expected free energy is naturally robust against occlusions, as it will actively explore the environment when the target object is initially not in view (Section~\ref{sect:results}).
   
\end{itemize}

\section{Related Work}

\textbf{Object-centric representations:} One of the early works on decomposing images of scenes into their constituent objects is Attend Infer Repeat~\citep{eslami_attend_2016}, where an observation is decomposed into object-level representations by having a recurrent neural network predicting the parameters of a spatial transformer network~\citep{jaderberg_spatial_2015}. PSGNets~\citep{Bear2020} introduce graph pooling and vectorization operations that convert image feature maps into object-centric graph structures. In ROOTS~\citep{chen_roots_2021}, the authors consider multiple viewpoints from the same scene, crop out the objects of each viewpoint, and then group them together to encode them with a generative query network instance for each object~\citep{eslami_neural_2018}. In doing so, the objects can be rendered individually and aggregated in a full observation. Other methods adopt a representation with a fixed number of ``slots'' that can be used to represent objects in the scene. MONet~\citep{burgess_monet_2019} recurrently predicts a mask for each available slot and then uses a variational autoencoder to encode each masked observation. In IODINE~\citep{greff_multi-object_2020}, a joint decomposition and representation model is learned with fixed slot allocation. To scale this up to more objects, the encoders are adapted for predicting proposals in parallel instead of recurrently~\citep{crawford_exploiting_2019,JiangJanghorbaniDeMeloAhn2020SCALOR}. GENESIS in addition learns an autoregressive prior for scene generation~\cite{engelcke2020a}. Locatello et al. introduce the Slot Attention module, which uses an attention mechanism to bind input features to the set of slots, which proved to be more efficient in terms of both memory consumption and runtime~\citep{Locatello2020}. 
Object Scene Representation Transformer~\citep{Mehdi2022} combines slot attention with scene representation transformers~\citep{srt22} to learn object-centric representations from a set of multiple viewpoints of a scene in an end-to-end fashion.
When dealing with sequences of observations over time, most models learn to predict the dynamics of each object slot separately~\citep{kosiorek_sequential_2018,JiangJanghorbaniDeMeloAhn2020SCALOR}.

\textbf{World Models:} In the context of reinforcement learning, world models have been devised to compress observations into a latent state space, and learn a dynamics model to predict how actions evolve future states~\cite{worldmodels}. These world models can then be used for planning actions~\cite{planet}, or for learning policies in imagination, achieving state-of-the-art performance on various RL benchmarks~\citep{Hafner2020Dream,Dreamerv2,Mazzaglia2022Choreographer}. An increasingly popular paradigm, especially in RL for robotics, is to provide the agent with a goal observation to obtained~\citep{HER}, for which also a world model can be leveraged to discover and achieve novel goals~\citep{lexa}. A recent line of work tries to combine structured latent state spaces, i.e. using different slots, for learning world models, typically for object-related tasks such as block pushing and stacking~\citep{kipf2019contrastive,Watters2019,Veerapaneni20a,Lin2020}.

\textbf{Active Inference:} Active inference is a theory that characterizes perception, planning, and action in terms of probabilistic inference~\citep{parr_active_2022}, which is applied in various domains ranging from neuroscience~\citep{Smith2020}, neuropsychology~\citep{Parr2018}, biology~\citep{Levin2022} to robotics and artificial intelligence~\citep{AIFroboticsSurvey}. In particular, visual foraging can be modeled using active inference agents, although typically limited to simulations with discrete observation or action spaces~\citep{mirza_scene_2016,dauce_active_2018}. 
\citet{van_de_maele_embodied_2022} adopt the active inference framework to learn object-centric generative models of 3D objects, called cortical column networks (CCN). Similar to the work on structured world models, \citet{vanbergenobai} extend the IODINE framework with actions to infer policies in a 2D sprites environment.

\textbf{Scene Representation Benchmark:} There exists a multitude of datasets for learning 3D scene representations, but many of these datasets use a static camera ~\cite{clevr,clevr3d,kubric} or use prerecorded moving camera frames~\cite{srt22}. To the best of our knowledge, none of these benchmarks consider an active camera that can move in the scene.

\section{Method}
\label{sect:method}

In this section, we describe how our agent carries out scene decomposition leveraging a hierarchical generative model. We first discuss the active inference framework and show how optimizing the free energy functional yields a trade-off between epistemic foraging and goal-directed behavior. Then, we propose a generative model representing a scene as a collection of multiple objects with distinct features, represented as hidden variables. Next, we describe how to instantiate the generative model using deep neural networks, and how the agent can be driven towards goal through expected free energy minimization, updating its beliefs about the scene representation at every step.

\subsection{Active Inference}

Active inference is a process theory of the brain which states that all neuronal processing and action is driven by the minimization of (a bound on) surprise, i.e. free energy~\cite{parr_active_2022}. This offers a first principles account of understanding perception and action as approximate Bayesian inference on hidden states and actions of a generative model. In general, this generative model is the joint probability distribution over sequences of observations $\tilde{\mathbf{o}}$, actions $\tilde{\mathbf{a}}$ and hidden states $\tilde{\mathbf{s}}$:
\begin{equation}
P(\tilde{\mathbf{o}}, \tilde{\mathbf{a}}, \tilde{\mathbf{s}}) = \prod_{t} P(\mathbf{o}_t | \mathbf{s}_t ) P(\mathbf{s}_t | \mathbf{s}_{t-1}, \mathbf{a}_{t-1}) P(\mathbf{a}_{t-1})
\end{equation}
The goal of the agent is then to minimize the variational free energy $F$~\citep{parr_active_2022}, i.e. the negative evidence lower bound ~\citep{rezende_stochastic_2014,kingma_auto-encoding_2014}, by introducing the approximate posterior $Q(\mathbf{s}_t | \mathbf{o}_t)$.
\begin{equation}
\begin{split}
F &= \sum_t 
  D_{KL} [Q(\mathbf{s}_t | \mathbf{o}_t)||  P(\mathbf{s}_t | \mathbf{s}_{t-1}, \mathbf{a}_{t-1})] \\
  &\quad+ \mathbb{E}_{Q(\mathbf{s}_t | \mathbf{o}_t)}[ - \log P(\mathbf{o}_t | \mathbf{s}_t )]
\end{split}
\end{equation}
Crucially, in active inference the agent also selects actions $a_t$ that it believes will minimize the so-called expected free energy $G(a_t)$~\citep{parr_active_2022}:
\begin{equation}
\begin{split}
G(a_t) &= \mathbb{E}_{Q(\mathbf{s}_{t+1}, \mathbf{o}_{t+1})}[ \underbrace{- \log P(\mathbf{o}_{t+1})}_{\text{Realizing Preferences}} \\
  &\quad+ \underbrace{\log Q(\mathbf{s}_{t+1} | \mathbf{a}_{t}) - \log Q(\mathbf{s}_{t+1} | \mathbf{a}_{t}, \mathbf{o}_{t+1})}_{\text{Expected Information Gain}}],
\end{split}
\label{eq:efe}
\end{equation}
consisting of a utility term based on a preferred distribution of future observations, as well as an information gain term. Hence, minimizing expected free energy balances goal-directed behavior and epistemic foraging.

\subsection{A Generative Model for Vision}
\label{sect:generativemodel}

\begin{figure}[t!]
    \centering
    \includegraphics[width=0.48\textwidth]{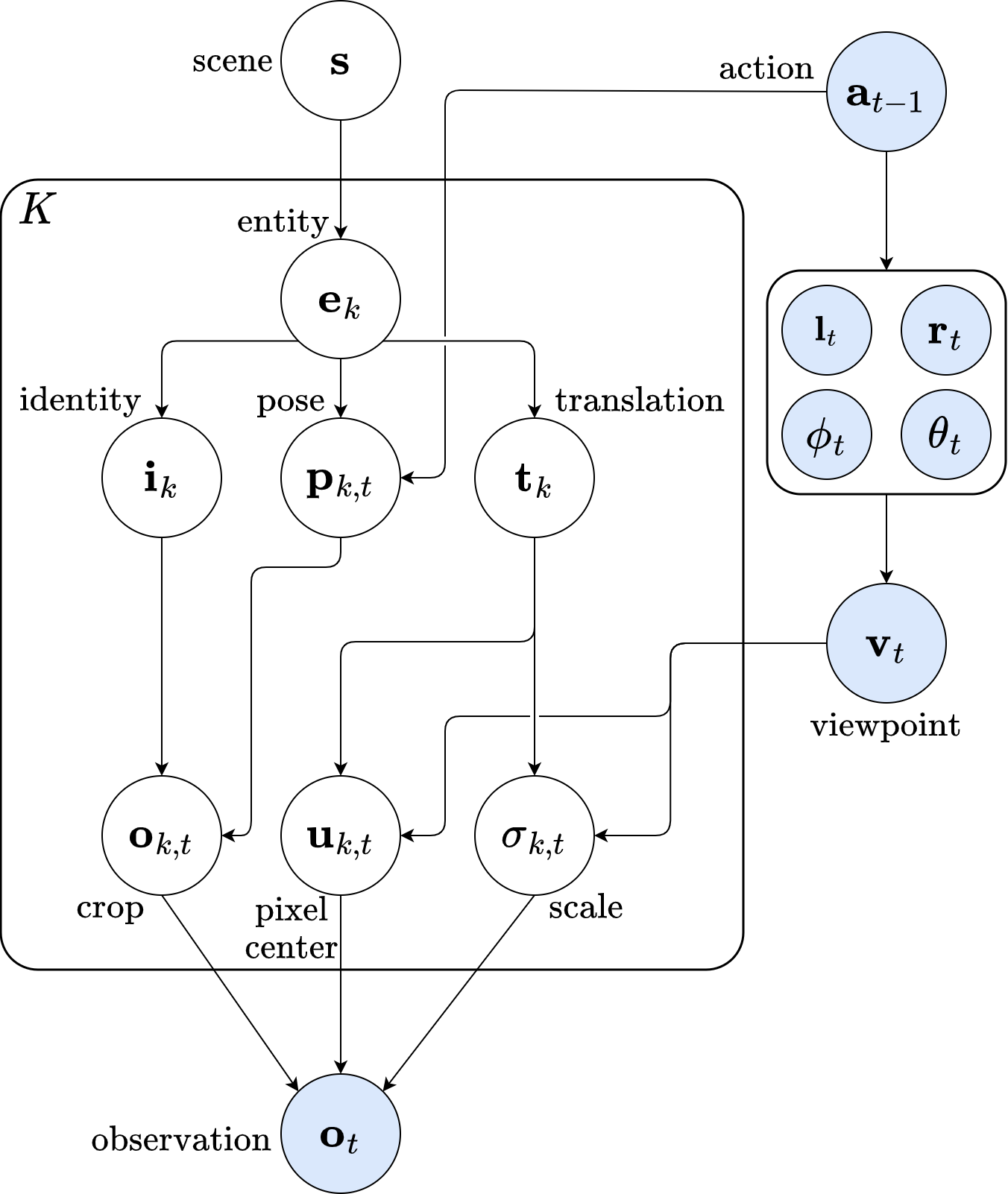}
    \caption{Graphical representation of the agents' generative model for an object-centric factorization of a scene. Observed variables are denoted by blue circles, while unobserved variables are denoted by white circles. }
    \label{fig:generativemodel}
\end{figure}

The agent - an active camera - entails a generative model to explain its observations and the effect of its actions thereon. In this case, the model should thus describe a 3D scene consisting of a set of objects in a workspace while observing 2D images from viewpoints reached through moving the camera. We factorize the generative model of a scene $\mathbf{s}$ as a hierarchical composition of $K$ entities $\mathbf{e}_k$. In the remainder, we assume the scene is static and therefore these entities do not change. However, this model could also be extended to take the object dynamics into account over time. 

The image the agent observes is constructed through a top-down generative model, depicted in Figure~\ref{fig:generativemodel}. Each entity $\mathbf{e}_k$ consists of an identity $\mathbf{i}_k$, a translation $\mathbf{t}_k$ w.r.t. a global reference frame, and a latent representation of the object pose $\mathbf{p}_{k,t}$. 

To compose an observation of a particular viewpoint $\mathbf{v}_t$, for each entity, an object-centric observation $\mathbf{o}_{k,t}$ is generated from the identity and pose, which renders the specific object in a particular pose. Next, given the object translation and a camera pinhole model, the pixel-coordinates $\mathbf{u}_{k,t}$ and scale $\mathbf{\sigma}_{k,t}$ in the full observation can be generated. The resulting observation $\mathbf{o}_{t}$ the agent receives as input is then a composition of these object-centric observations in a global view. Note that the identity and translation parameters are consistent over the scene, do not change over time, and are therefore not dependent on the action of the agent. 

The action $\mathbf{a}_{t-1}$ represents the relative transform the agent must take in the global reference frame, in order to reach the next viewpoint $\mathbf{v}_{t}$. We parameterize the resulting viewpoint as the combination of the location $\mathbf{l}_t$ in space at which the agent wants to direct its gaze, together with the spherical coordinates with respect to this look-at point: range $\mathbf{r}_t$, elevation $\mathbf{\phi}_t$ and azimuth $\mathbf{\theta}_t$.

\begin{figure*}[t!]
    \centering
    \includegraphics[width=\textwidth]{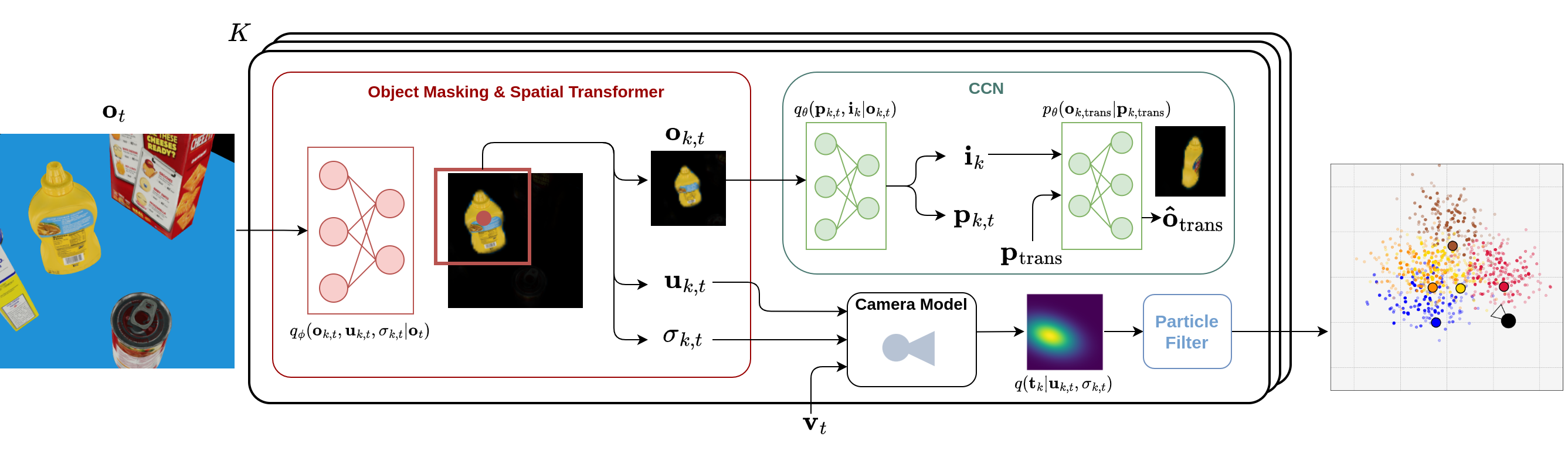}
    \caption{Flow of the inference process of observation into the distinct latent variables. For each of the K-considered object categories, the object pixels are extracted from the observation (red) by a neural network that predicts pixel-center, scale, and a masked crop of the object. From this crop, the belief over the pose and identity latent variable is predicted (green) using a CCN. Given the pinhole camera model, pixel-wise center, and scale of the object, an estimate of the object's location are formed which is used to update a particle filter (blue) approximating the belief over the object translation.}
    \label{fig:nn_flow}
\end{figure*}

The factorization of the described generative model is shown in Figure~\ref{fig:generativemodel}, and a detailed formal description of the model is provided in Appendix~\ref{appendix:fe}. This kind of decomposition of a scene in distinct objects with their respective features, as well as the specification of actions in our generative model might also underpin how the visual system of the brain works, and maps to the visual cortices, cerebral attention networks, and the oculomotor system~\citep{parr_generative_2021}.

\subsection{Scene Perception with SceneCCN}
\label{sect:perception} 

In active inference, perception is cast as approximate Bayesian inference over the hidden variables. At each timestep, the agent observes $\mathbf{o}_t$ and infers approximate posterior distributions over each hidden variable in a bottom-up process. This inference process is amortized using a collection of neural network models, which is depicted in Figure~\ref{fig:nn_flow}.

We ground our approach on Cortical Column Networks (CCN)~\citep{van_de_maele_embodied_2022}, which are object-centric models that learn the approximate posterior over the latent pose $q_\theta(\mathbf{p}_{k,t}|\mathbf{o}_{k,t})$ for views of a particular object, by predicting observations of novel poses. These consist of a separate encoder-decoder model trained separately per object category, which is inspired by cortical columns in the brain~\citep{hawkins_theory_2017}, and reflect how toddlers typically interact with one object at a time~\citep{Karin2014}. This allows these to be efficiently trained in parallel on a limited dataset consisting of views of one particular object, at the cost of poor generalization (i.e. for new object categories a new CCN will need to be trained). In addition to the pose representation, the encoder also outputs a Bernoulli variable $q_\theta(\mathbf{i}_{k}|\mathbf{o}_{k,t})$ which indicates the belief that the object of interest is present in the view.

In order to infer object-centric views $\mathbf{o}_{k,t}$, we learn for each object category an object extraction network $q_\phi(\mathbf{o}_{k,t},\mathbf{u}_{k,t},\mathbf{\sigma}_{k,t}|\mathbf{o}_t)$ with parameters $\phi$, which first produces a mask $\mathbf{\alpha}_{k,t}$ which masks anything beside the object category of interest. Based on the masked observation $\mathbf{o}_t\odot\mathbf{\alpha}_{k,t}$, we then predict the center pixel $\mathbf{u}_{k,t}$ of where the object is present in the observation, as well as the scale $\mathbf{\sigma}_{k,t}$. This is then fed into a spatial transformer network~\citep{jaderberg_spatial_2015}, which produces an object-centric crop $\mathbf{o}_{k,t}$ for the CCN.

The neural networks comprising these two blocks are optimized in two separate phases. In the first phase, the pose representation is learned by training the CCN on an object-centric dataset. For each object category, a set of 500 observations and viewpoint pairs is collected. The viewpoints are sampled uniformly on the surface of a sphere, with a fixed radius, while oriented towards the object center. For each step during training, two views are randomly sampled, and the action is computed as the relative displacement in azimuth and elevation. To learn how this action results in a transitioned pose representation, we also train a pose transition network $p_\theta(\mathbf{p}_{k,2}|\mathbf{p}_{k,1}, \mathbf{a}_{t-1})$ that models these dynamics. For more details about the construction of this dataset, the reader is referred to Appendix~\ref{appendix:dataset}. 

The objective is prediction error over transitioned poses, given an action:
\begin{align}
    \label{eq:l1}
        \mathcal{L}_1 &= \underbrace{\lambda_1 || \mathbf{o}_{k,1} - p_\theta(\mathbf{o}_{k,1}|\mathbf{p}_{k,1}) ||_2 
        + \lambda_2 || \mathbf{o}_{k,2} - p_\theta(\mathbf{o}_{k,1}|\mathbf{p}_{k,1}) ||_2}_{\text{Reconstruction error before transition}} \nonumber \\ 
        &\quad+ \underbrace{\lambda_3\  || \mathbf{o}_{k,2} - p_\theta({\mathbf{o}}_{k,2,\text{trans}}|\mathbf{p}_{k,2,\text{trans}})||_2}_{\text{Reconstruction error after transition}} \\
        &\quad+ \underbrace{\lambda_4\  D_{KL}[q_\theta(\mathbf{p}_{k,2}|\mathbf{o}_{k,2})||p_\theta(\mathbf{p}_{k,2}|\mathbf{p}_{k,1}, \mathbf{a}_{t-1})]}_{\text{Complexity of transition model}}, \nonumber
\end{align}
These models are additionally regularized by a KL-divergence term between the outputted distributions and a standard normal distribution. 

In the second phase, the object extraction network is trained for extracting object-centric crops. To this end, we do not collect a new dataset, but instead augment the object-centric dataset by scaling and translating the observations, while also adding a scaled and translated object from a different category to simulate occlusion. On top of this, we add random color patches in the background. The model learns to extract object-centric crops by minimizing:
\begin{equation}
    \label{eq:l2}
    \begin{split}
        \mathcal{L}_2 &= \underbrace{||\mathbf{o}_{k,t}-q_\phi(\mathbf{o}_{k,t}|\mathbf{u}_{k,t},\mathbf{\sigma}_{k,t},\mathbf{o}_t)||_2}_{\text{Cropping error}} \\
        &\quad+ \underbrace{||\mathbf{\alpha}_{k,t} - \mathbf{\hat{\alpha}}_{k,t}||_2}_{\text{Masking error}} + \underbrace{\text{BCE}(\mathbf{i}_k, q_\theta(\mathbf{i}_k|\mathbf{o}_{k,t}))}_{\text{Identification error}},
    \end{split}
\end{equation}

where the first two terms are computed as the mean squared error between the predicted masks and crops, and the respective ground truths. The BCE-term is the binary cross entropy loss over the predicted object identity variable. For regularization purposes, the extracted crop is also fed through the (frozen) CCN and the KL-divergence terms from Equation~\ref{eq:l1} are also optimized.

In Appendix~\ref{appendix:fe}, we show how these loss terms are in effect consistent with minimizing the variational free energy of the generative model proposed Section~\ref{sect:generativemodel}. Also for the exact parameterizations of all neural networks, hyperparameters, and training details, the reader is referred to Appendix~\ref{appendix:nn}. 

To infer posterior beliefs over the object translation $\mathbf{t}_{k,t}$ in an allocentric reference frame, we use the inferred pixel coordinates $\mathbf{u}_{k,t}$ and scale $\mathbf{\sigma}_{k,t}$ of the object. Given the absolute viewpoint of our camera and a pinhole camera model, we can backproject the pixel coordinate to a ray in 3D, along which we specify a 3D Gaussian distribution with mean at the estimated depth given the scale, and a fixed covariance matrix forming an ellipsoid density along the ray. To get a better estimate of the object's position over time, this position belief is integrated using a particle filter. When an object is not detected in an observation, we also reduce the particle weights that are in view, to also reflect and integrate the information gained by not observing a particular object.

\subsection{Action selection\label{sect:actions}}

Setting an agent's goal in active inference is done by specifying a prior preference in observation space, which the agent is expected to obtain. In this case, the preference is the log probability of reaching the goal observation $\mathbf{o}_\text{goal}$. Actions are then selected that minimize the expected free energy G, as defined in Equation~\ref{eq:efe}, i.e. scoring how much expected observations will realize the preferred observations, and how much information these will bring over the hidden variables. 

In our case, we mainly focus on the expected information gain about the object positions $\mathbf{t}_{k}$, which are the main source of uncertainty in the scene. Moreover, since the log likelihood in observation space (i.e. pixel space) is less meaningful, we first infer the goal object identity, scale and pose from $\mathbf{o}_\text{goal}$ using our SceneCCN, and then score utility w.r.t. reaching these factor values. 

The expected free energy for a candidate viewpoint $\mathbf{v}_{t+1}$ becomes:
\begin{align}
    \label{eq:g}
    G(\mathbf{v}_{t+1}) &= \mathbb{E}_{Q(\mathbf{s}, \mathbf{o}_{t+1})}\big[
    \underbrace{-\log P(\mathbf{t}_k | \mathbf{v}_{t+1}, \mathbf{o}_{\text{goal}})}_{\text{Look at goal object position}} \nonumber \\
    &\quad - \underbrace{\log P(\mathbf{\sigma}_{k,t+1} | \mathbf{v}_{t+1}, \mathbf{o}_{\text{goal}})}_{\text{at the desired scale}} \nonumber \\
    &\quad - \underbrace{\log P(\mathbf{p}_{k,t+1} | \mathbf{v}_{t+1}, \mathbf{o}_{\text{goal}})}_{\text{and desired pose}} \\
    &\quad + \underbrace{
    \log Q(\mathbf{t}_{k, t+1} | \mathbf{v}_{t+1}) 
    - \log Q(\mathbf{t}_{k,t+1} | \mathbf{v}_{t+1}, \mathbf{o}_{t+1})
    }_{\text{while searching for the object position.}}\big]  \nonumber
\end{align}

This boils down to directing the agent towards looking at the goal object at the right pose and scale, while also searching for where the object is positioned. We use Monte Carlo sampling to obtain the best next viewpoint, by sampling 5000 targets and evaluating G. Instead of sampling uniformly in the workspace, we use importance sampling, putting more weight on viewpoints that look at positions where the target object is more likely positioned given the current beliefs of the particle filter. Once $\mathbf{v}_{t+1}$ is determined, we find the next action, by moving the camera a step of maximum 5 cm towards the target view. 

\section{Experiments}

We aim to evaluate whether our proposed generative model entails a representation that enables the agent to understand the scene and to infer: \emph{(i)} what the different objects are and \emph{(ii)} where these different objects are located. To this end, we design a new environment in which objects from the YCB dataset~\cite{ycb} are spawned in random positions, and the agent can move a camera in the workspace.

\subsection{Active Search Benchmark}
\label{sect:env}

In order to evaluate an active vision agent, we designed a benchmark that casts the scene perception problem as a problem in which a goal observation has to be reached by a virtual camera. We created a simulation environment in which between one and five objects from the YCB dataset~\cite{ycb} are spawned. We consider the master chef can, the cracker box, the sugar box, the tomato soup can, and the mustard bottle. The positions of the objects are randomly generated to be on a table of size 1m x 1m with a uniformly randomly sampled color. The goal image is an object-centric observation of a target object. In this observation, the other objects in the scene are not rendered to have no ambiguity on what the target is. 

The agent can move a pinhole camera, and the action space is defined as continuous in 6 dimensions, representing the relative displacement in the three-dimensional space and the Euler angles for the relative rotation. Values in the action space lie in the range $[-0.5, 0.5]$. The agent can only move over the table with a max height of 0.6 cm. If the action moves the camera outside of the environment or inside of an object, the resulting pose is clipped.

\begin{figure}[t]
    \centering
    \includegraphics[width=0.48\textwidth]{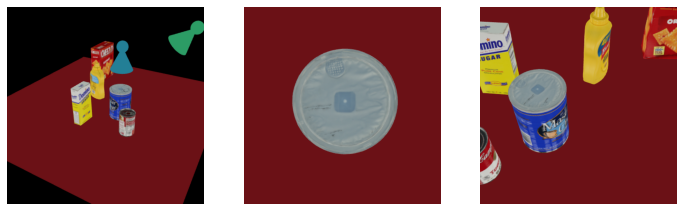}
    \caption{An example of the active vision environment. The left image shows the scene, a table with 5 objects from the YCB dataset. The green camera shows the current pose of the agent and the blue camera shows the target pose of the agent in the environment. The middle image shows the object-centric target goal observation in which the other objects are not rendered. The right image shows the current observation the agent sees.}
    \label{fig:environment}
\end{figure}

The task is considered successful if the agent reaches the goal pose within a translation error below $7.5$ cm, and a rotation error lower than $0.5$ rad. The agent starts at an initial position with $0$ rad azimuth, $\frac{\pi}{4}$ rad elevation, and a range of $0.65$ m, with respect to the table center. A visual representation of the environment, a goal observation, and the initial observation of a randomly generated scene is shown in Figure~\ref{fig:environment}.

We benchmark the performance in this environment using the following metrics: the success rate (i.e. when the agent reaches the goal in less than 350 steps), the azimuth, elevation, and range error of the goal camera position with respect to the goal object position. As symmetry in object shape can cause a high overall translation error while actually having an accurate final observation, opted for these metrics instead of an error in allocentric 6DOF space.

We implemented this environment adopting the structure of OpenAI gym~\cite{brockman2016openai}, and the code is supplied in the supplementary material.

\subsection{Results}
\label{sect:results}

\textbf{Baselines:} We compare our approach with several baselines for solving the benchmark: 
\begin{itemize}
    \item \textbf{SceneCCN + AIF}: Our agent uses the proposed SceneCCN to decompose scene observations into beliefs over separate factors (Section 3.3), and adopts the expected free energy objective for action selection (Section 3.4).
    \item \textbf{PoseCNN}: A supervised model for object segmentation and pose estimation of RGBD data~\cite{xiang_posecnn_2018} trained on the YCB dataset. The agent action is acquired by first estimating the identity and pose directly from both the goal observation and the agents' current observation. From these two estimated poses, the relative transform is computed, and a 5 cm step according to this transform is taken as action. The agent executes this process every step until the target pose is reached. When the agent does not observe the preference the agent stops. Note that PoseCNN has access to the depth information, whereas our approach has not.
    \item \textbf{PoseCNN + Infogain}: The PoseCNN baseline, but adding our object position particle filter. When the target object is not in view, we use the information gain term over object position similar to our approach.
    \item \textbf{LEXA}: A model-based RL agent that adopts unsupervised exploration, to learn an accurate world model of the environment, and goal-conditioned RL, to learn to reach goals sampled from the agent experience buffer~\cite{lexa}. 
\end{itemize}

\begin{figure}[b!]
    \centering
    \includegraphics[width=0.48\textwidth]{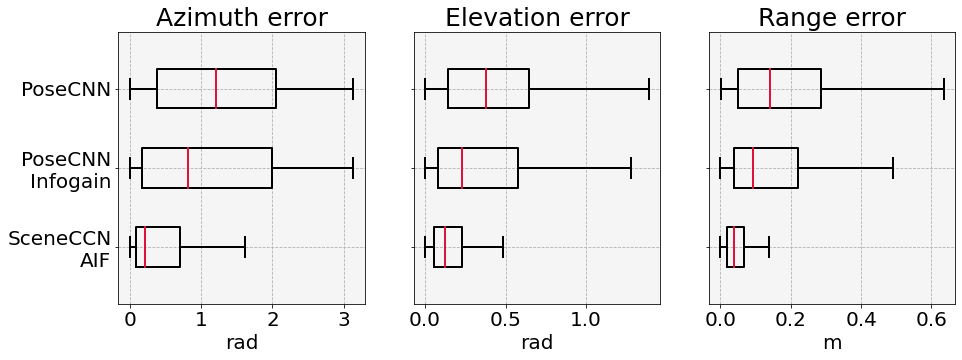}
    \caption{Box plot representing the azimuth, elevation, and range error of the reached position for our approach and the two PoseCNN baselines over all scenes of the 25 configurations.}
    \label{fig:benchmark_boxes}
\end{figure}

\begin{table*}[]
    \centering
    \caption{Comparison of SceneCCN with PoseCNN on the whole benchmark. We compare over the success rate (\textbf{\% s}), azimuth error $\Delta \mathbf{\phi}$, elevation error $\Delta \mathbf{\theta}$ and range error $\Delta \mathbf{r}$. For each target object, 100 scenes are evaluated. The values are represented as mean $\pm$ standard error. The best performances are marked in bold. } 
    \begin{tabular}{llcccc}
        \toprule[1.5pt]
        & & \textbf{\% s} & \textbf{$\Delta \mathbf{\phi}$} & \textbf{$\Delta \mathbf{\theta}$} & \textbf{$\Delta \mathbf{r}$} \\
        \midrule[1.5pt]
        \multirow{6}{*}{PoseCNN}  & Master chef can & $17.0$ & $1.333_{\pm 0.088}$ & $0.418_{\pm 0.029}$ & $0.203_{\pm 0.018}$\\
        & Cracker box & $24.0$ & $1.219_{\pm 0.096}$ & $0.387_{\pm 0.031}$ & $0.148_{\pm 0.013}$\\
        & Sugar box & $16.0$ & $1.092_{\pm 0.087}$ & $0.447_{\pm 0.035}$ & $0.185_{\pm 0.015}$\\
        & Mustard bottle & $8.0$ & $1.466_{\pm 0.091}$ & $0.508_{\pm 0.041}$ & $0.175_{\pm 0.016}$\\
        & Tomato soup can & $16.0$ & $1.276_{\pm 0.100}$ & $0.395_{\pm 0.030}$ & $0.218_{\pm 0.018}$\\
        & Total  & $16.2$ & $1.277_{\pm 0.042}$ & $0.431_{\pm 0.015}$ & $0.186_{\pm 0.007}$ \\
        \midrule
        \multirow{6}{*}{PoseCNN + Infogain}  & Master chef can & $22.0$ & $1.198_{\pm 0.099}$ & $0.419_{\pm 0.034}$ & $0.171_{\pm 0.017}$\\
        & Cracker box & $28.0$ & $1.115_{\pm 0.101}$ & $0.380_{\pm 0.036}$ & $0.133_{\pm 0.011}$\\
        & Sugar box & $35.0$ & $0.978_{\pm 0.098}$ & $0.330_{\pm 0.032}$ & $0.149_{\pm 0.015}$\\
        & Mustard bottle & $24.0$ & $1.241_{\pm 0.100}$ & $0.381_{\pm 0.037}$ & $0.138_{\pm 0.013}$\\
        & Tomato soup can & $33.0$ & $1.065_{\pm 0.097}$ & $0.327_{\pm 0.031}$ & $0.157_{\pm 0.018}$\\
        & Total  & $28.4$ & $1.119_{\pm 0.045}$ & $0.367_{\pm 0.015}$ & $0.150_{\pm 0.007}$ \\
        \midrule
        \multirow{6}{*}{SceneCCN + AIF}  & Master chef can & $\mathbf{84.0}$ & $\mathbf{0.491}_{\pm 0.067}$ & $\mathbf{0.138}_{\pm 0.014}$ & $\mathbf{0.042}_{\pm 0.005}$\\
        & Cracker box & $\mathbf{66.0}$ & $\mathbf{0.532}_{\pm 0.075}$ & $\mathbf{0.290}_{\pm 0.039}$ & $\mathbf{0.058}_{\pm 0.007}$\\
        & Sugar box & $\mathbf{69.0}$ & $\mathbf{0.508}_{\pm 0.076}$ & $\mathbf{0.232}_{\pm 0.032}$ & $\mathbf{0.058}_{\pm 0.006}$\\
        & Mustard bottle & $\mathbf{80.0}$ & $\mathbf{0.420}_{\pm 0.064}$ & $\mathbf{0.275}_{\pm 0.041}$ & $\mathbf{0.057}_{\pm 0.008}$\\
        & Tomato soup can & $\mathbf{46.0}$ & $\mathbf{0.896}_{\pm 0.085}$ & $\mathbf{0.220}_{\pm 0.019}$ & $\mathbf{0.120}_{\pm 0.013}$\\
        & Total  & $\mathbf{69.0}$ & $\mathbf{0.569}_{\pm 0.034}$ & $\mathbf{0.231}_{\pm 0.014}$ & $\mathbf{0.067}_{\pm 0.004}$ \\
        \bottomrule[1.5pt]
    \end{tabular}
    \label{tab:benchmark_table}
\end{table*}

\textbf{Benchmark Performance:} The benchmark consists of 500 evaluation scenes: 100 for each target object with 1-5 objects randomly positioned in the workspace, which has a random color.
The exact scenes on which we evaluated are provided in the supplementary material. The error in spherical coordinates with respect to the target object center is plotted in Figure~\ref{fig:benchmark_boxes}. Our approach outperforms the PoseCNN baseline, whereas adding information gain also improves the vanilla PoseCNN baseline.

A more detailed breakdown per object is provided in Table~\ref{tab:benchmark_table}. Our model consistently outperforms the baselines in both in terms of pose error and success rate of reaching the target pose according to the environment stop criteria. We obtain an average success rate of 69\%, which is more than double than the PoseCNN models with and without info gain (28.4\% and 16.2\% respectively). This shows that using an infogain exploration strategy clearly helps in finding the target object, but the object-centric model is also an important success factor. This acknowledges that representations learnt by actively interacting with objects might be better suited for object understanding than supervised training of 6DOF pose regression on static views.  

When we disentangle these features separately for the different objects, shown in Table~\ref{tab:benchmark_table}, we observe that the azimuth error is the highest for the object with the most symmetry (i.e. tomato soup can), and the lowest for the object with the least symmetry (i.e. the mustard bottle), while this is not necessarily reflected by the elevation. We also to note that the performance of our SceneCCN on the tomato soup can underperform compared to the other objects. We attribute this to the fact that the CCN model of the tomato soup can was less accurate than the ones of the other objects. This is probably due to the fact that the tomato soup can was rendered in a smaller scale than the others (see Figure~\ref{fig:dataset}), which made it harder for the models to accurately infer the position and pose.

\begin{figure}[b!]
    \centering
    \includegraphics[width=0.48\textwidth]{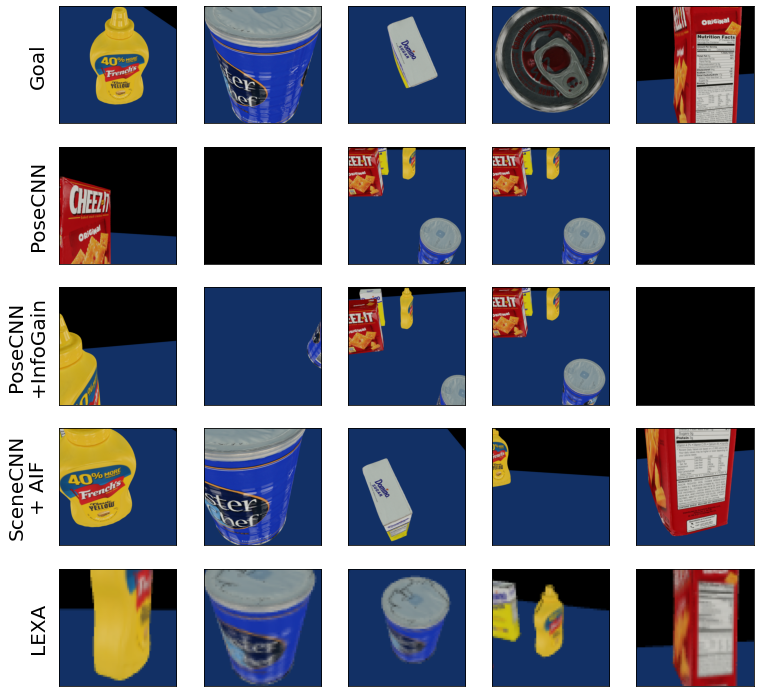}
    \caption{Five goals in the environment on which the LEXA agent was trained. We show the finally reached frame for all the baselines (PoseCNN, PoseCNN + Infogain, SceneCCN + AIF and LEXA).}
    \label{fig:qualitative_lexa}
\end{figure}

\textbf{Comparison with model-based RL:} We also evaluated Latent Explorer Achiever (LEXA)~\cite{lexa} on a single environment instance with 5 objects in a fixed configuration, but with varying goals to reflect the original LEXA setup. LEXA sets goals by randomly sampling them from a replay buffer acquired through exploration. We trained this agent for $\sim$7M steps\footnote{For sample-efficiency comparison, we trained SceneCCN using 500k samples per object, for a total of 2.5M frames on the augmented dataset.} and compare our approach and both PoseCNN approaches on this environment. Qualitative results are shown in Figure~\ref{fig:qualitative_lexa} and show the goal observation, and the final reached frame for all the baselines as well as for our approach. We can see that the LEXA agent sometimes is able to reach the goal almost perfectly (second and last column) but in other cases, it picks the wrong target object. 

The quantitative performance of all approaches in this environment is shown in Table~\ref{tab:lexatable}. We observe that the LEXA baseline is only able to solve the task in 11 out of 100 cases, while our approach was able to solve this 62 times. PoseCNN was only able to  solve it 17 times and 30 times with the added infogain term. Further, we notice the same trends as observed in Table~\ref{tab:benchmark_table}. While LEXA performs worse than the PoseCNN baselines in terms of success rate, the range and elevation error are lower. It also has the lowest azimuth error of all approaches. 

\begin{table}[h]
    \centering
    \caption{Comparison with LEXA on the single environment LEXA was trained on. 100 goals were randomly selected and used to evaluate the performance of the agents on the success rate (\textbf{\% s}), azimuth error ($\Delta \mathbf{\phi}$), elevation error ($\Delta \mathbf{\theta}$) and range error ($\Delta \mathbf{r}$).}
    \label{tab:lexatable}
    \begin{tabular}{lcp{8mm}p{8mm}p{8mm}}
        \toprule[1.5pt]
        & \textbf{\% s} & \textbf{$\Delta \mathbf{\phi}$} & \textbf{$\Delta \mathbf{\theta}$} & \textbf{$\Delta \mathbf{r}$} \\
        \midrule[1.5pt]
        PoseCNN & $17$ & $1.824$ $^{\pm0.494}$ & $0.425$ $^{\pm0.082}$ & $0.238$ $^{\pm0.053}$\\
        PoseCNN + Infogain & $30$ & $1.195$ $^{\pm0.398}$ & $0.460$ $^{\pm0.118}$ & $0.201$ $^{\pm0.057}$\\
        SceneCCN + AIF & $\mathbf{62}$ & $1.050$ $^{\pm0.470}$ & $\mathbf{0.187}$ $^{\pm0.083}$ & $\mathbf{0.076}$ $^{\pm0.039}$\\
        LEXA & $11$ & $\mathbf{1.025}$ $^{\pm0.300}$ & $0.368$ $^{\pm0.103}$ & $0.138$ $^{\pm0.030}$\\
        \bottomrule[1.5pt]
    \end{tabular}
\end{table}

\begin{figure}[t]
    \centering
    \includegraphics[width=0.48\textwidth]{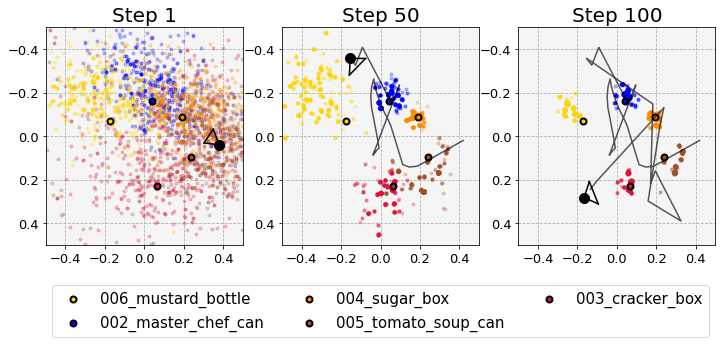}
    \caption{2D projection of the position belief evolution through the minimizing of the expected free energy $G$ using the SceneCCN generative model. The colors indicate the object over which the belief is formed. The black circles represent the ground truth positions of these objects and the black line shows the trajectory taken by the agent.}
    \label{fig:position}
\end{figure}

\textbf{Exploration vs Exploitation:} Finally. we investigate the emerging behavior when the agent does not have a goal image. The function optimized by the agent then becomes the expected free energy without the instrumental terms, which boils down to the (negative) infogain over the object position (see Equation~\eqref{eq:g}). In Figure~\ref{fig:position}, we show this for a particular scene in which the five YCB objects are present. At the first step, it is clear that the belief over object positions has a large variance, but over time the agent iterates and attends to different objects, narrowing the variance over these specific object instances. This can be seen by the camera's trajectory shown in black. We observe that over multiple timesteps the variance over the object position reduces, and the particles mean lies closer to the ground truth position of the objects. In this visualization, we show that driving behavior through the minimization of G actively reduces the ambiguity over the scene.

\section{Conclusion}

In this paper, we proposed a novel object-centric generative model that builds a representation for scenes with multiple objects using active inference. We developed a new benchmark for evaluating performance on active vision tasks using a pinhole camera agent in a 3D environment with a specific move-to goal. Our approach outperforms both a supervised pose estimation model and an unsupervised model-based RL baseline.

The main limitation of SceneCCN is that we require a single neural network per object type, which will be difficult to scale to a world with hundreds of object categories with even more different appearances. A potential mitigation would be to extend our CCN models to share weights of the different encoders, and/or also infer a latent representation for shape and appearance as proposed by~\citet{ferraro_disentangling}. 

Compared to slot-attention models, we entail a separate slot per object entity instead of a slot per object instance. This ensures that we require less data to train a single category compared to an end-to-end system that has to learn to map either of the categories to any slot. However, our approach comes the downside that it is more difficult to scale towards scenes with multiple objects of a single instance, or to deal with heavily cluttered environments of similar categories.

Currently, our method only considers static environments. In future work, we plan to also consider object dynamics by adding a dynamics model for each object category that given the history of poses and action could predict the next pose.

\bibliography{references,references_new}
\bibliographystyle{icml2023}

\newpage
\appendix
\onecolumn
\section{The Generative Model and Variational Free Energy}
\label{appendix:fe}

The generative model shown in Figure~\ref{fig:generativemodel} can be formalized as: 

\begin{equation}
\begin{split}
    P(\mathbf{s},\Tilde{\mathbf{o}},\Tilde{\mathbf{a}}) &= P(\mathbf{v}_0) \prod_{t=1}^T 
     P(\mathbf{v}_t|\mathbf{v}_{t-1},\mathbf{a}_{t-1}) P(\mathbf{a}_{t-1}) \\
    &\quad \prod_{k} P(\mathbf{o}_t|\mathbf{o}_{k,t},\mathbf{u}_{k,t},\mathbf{\sigma}_{k,t}) P(\mathbf{o}_{k,t}|\mathbf{i}_k,\mathbf{p}_{k,t})  P(\mathbf{p}_{k,t}|\mathbf{p}_{k,t-1},\mathbf{a}_{t-1})
    P(\mathbf{u}_{k,t}|\mathbf{t}_k,\mathbf{v}_{t})
    P(\mathbf{\sigma}_{k,t}|\mathbf{t}_k,\mathbf{v}_t)
    P(\mathbf{i}_k) P(\mathbf{t}_k)
\end{split}
\end{equation}

where the tilde represents a sequence of the variable over time, and $P(\mathbf{s})=\prod_{k} P(\mathbf{i}_k, \mathbf{t}_k, \Tilde{\mathbf{p}}_k)$.

The approximate posterior $Q(\mathbf{s}|\Tilde{\mathbf{o}})$ is factorized through the following mean field approximation: 
\begin{equation}
    \begin{split}
    Q(\mathbf{s}|\Tilde{\mathbf{o}}) &= \prod_t \prod_k 
    Q(\mathbf{u}_{k,t}|\mathbf{o}_t)
    Q(\mathbf{\sigma}_{k,t}|\mathbf{o}_t)
    Q(\mathbf{t}_{k}|\mathbf{u}_{k,t},\mathbf{\sigma}_{k,t})
    Q(\mathbf{p}_{k,t}|\mathbf{o}_{k,t})
    Q(\mathbf{i}_k|\mathbf{o}_{k,t}) 
    \end{split}
\end{equation}

For the generative model and approximate posterior described above, the free energy $F$ is defined as: 

    \begin{equation}
        \label{eq:fe}
        \begin{split}
            F &= \mathbb{E}_{Q(\mathbf{s}|\Tilde{\mathbf{o}})} \big[\log Q(\mathbf{s}|\Tilde{\mathbf{o}}) - \log P(\mathbf{s},\Tilde{\mathbf{o}},\tilde{\mathbf{a}}) \big] \\     
            &= \sum_t \sum_k  \ \underbrace{\mathbb{E}_{Q(\mathbf{s}|\Tilde{\mathbf{o}})} \big[- \log P(\mathbf{v}_t|\mathbf{v}_{t-1},\mathbf{a}_{t-1}) - \log P(\mathbf{a}_{t-1}) \big]}_{\text{Constant}}\\
            &\quad\quad\quad\quad\quad + \underbrace{D_{KL} \big[Q(\mathbf{p}_{k,t}|\mathbf{o}_{k,t})||P(\mathbf{p}_{k,t}|\mathbf{p}_{k,t-1}, \mathbf{a}_{t-1}))\big]}_{\text{Pose Transition Model}^{*}} \\
            &\quad\quad\quad\quad\quad + \underbrace{D_{KL} \big[Q(\mathbf{i}_k|\mathbf{o}_{k,t})||P(\mathbf{i}_k)\big]}_{
            \text{Object category classification}^{**}
            } \\
            &\quad\quad\quad\quad\quad + \underbrace{D_{KL} \big[Q(\mathbf{\sigma}_{k,t}|\mathbf{o}_{t})||P(\mathbf{\sigma}_{k,t}|\mathbf{t}_{t}, \mathbf{v}_t)\big]
            + D_{KL} \big[Q(\mathbf{u}_{k,t}|\mathbf{o}_{t})||P(\mathbf{u}_{k,t}|\mathbf{t}_{t}, \mathbf{v}_t)\big]}_{\text{Object-centric cropping}^{**}} \\
            &\quad\quad\quad\quad\quad + \underbrace{D_{KL} \big[Q(\mathbf{t}_{k}|\mathbf{o}_{t})||P(\mathbf{t}_k)\big]}_{\text{Object positions}^{***}} \\
            &\quad\quad\quad\quad\quad +\underbrace{\mathbb{E}_{Q(\mathbf{s}|\Tilde{\mathbf{o}})}\big[-\log P(\mathbf{o}_{k,t}|\mathbf{i}_{t}, \mathbf{p}_{k,t})) \big]}_{\text{Object-centric observation likelihood}^{*}} \\
            &\quad\quad\quad\quad\quad + \underbrace{\mathbb{E}_{Q(\mathbf{s}|\Tilde{\mathbf{o}})} \big[-\log P(\mathbf{o}_t | \mathbf{o}_{k,t},\mathbf{u}_{k,t}, \mathbf{\sigma}_{k,t}) \big]}_{\text{Full observation likelihood}}\\
        \end{split}
    \end{equation}

This decomposes in a number of terms that are optimized in our two training phases. In the first phase, we train object-centric CCNs, which jointly optimize an object-centric likelihood model $p_\theta(\mathbf{o}_{k,t}|\mathbf{i}_{k}, \mathbf{p}_{k,t})$, a pose transition model $p_\theta(\mathbf{p}_{k,t}|\mathbf{p}_{k,t-1}, \mathbf{a}_{t-1})$ and an approximate pose posterior model $q_\phi(\mathbf{p}_{k,t}|\mathbf{o}_{k,t})$ covering the terms denoted with $^{*}$. In the second phase, we additionally train a model outputting the object category $q_\phi(\mathbf{i}_k|\mathbf{o}_{k,t})$ as well as a model outputting object-centric crop parameters $q_\phi(\mathbf{u}_{k,t}, \sigma_{k,t}|\mathbf{o}_{t})$. This is done by optimizing the terms denoted with $^{**}$, and noting that generating bounding box parameters $\mathbf{u}_{k,t}$ and $\sigma_{k,t}$ from object position $\mathbf{t}$ and camera viewpoint $\mathbf{v}$ is determined by the pinhole camera model.  Finally, the posterior over positions (denoted with $^{***}$) is implemented using a particle filter, and has no parameters to optimize. In a similar vein, the remaining terms have no influence over the learnable parameters or are constant terms under the expectation.

In practice, optimizing the negative log likelihood terms is equivalent to minimizing the mean squared error between the reconstructed observation and the ground truth observation. Similarly, optimizing the KL-divergence for a Bernoulli variable is equivalent to minimizing the binary cross entropy error over this variable. For this reason, we implemented the loss using these terms. Instead of regressing the bounding box parameters, we opted for a spatial transformer that directly transforms the input image into an object-centric crop from these parameters, and add a supervised loss to the intermediately computed object-specific mask as this resulted in more stable results.


\section{Object-Centric Datasets}
\label{appendix:dataset}

For each of the considered objects in the YCB dataset, a dataset is created consisting of observation-viewpoint pairs. In this work we consider the following objects: master chef can, cracker box, sugar box, tomato soup can, and mustard bottle. 

\begin{figure}
    \centering
    \includegraphics[width=0.8\textwidth]{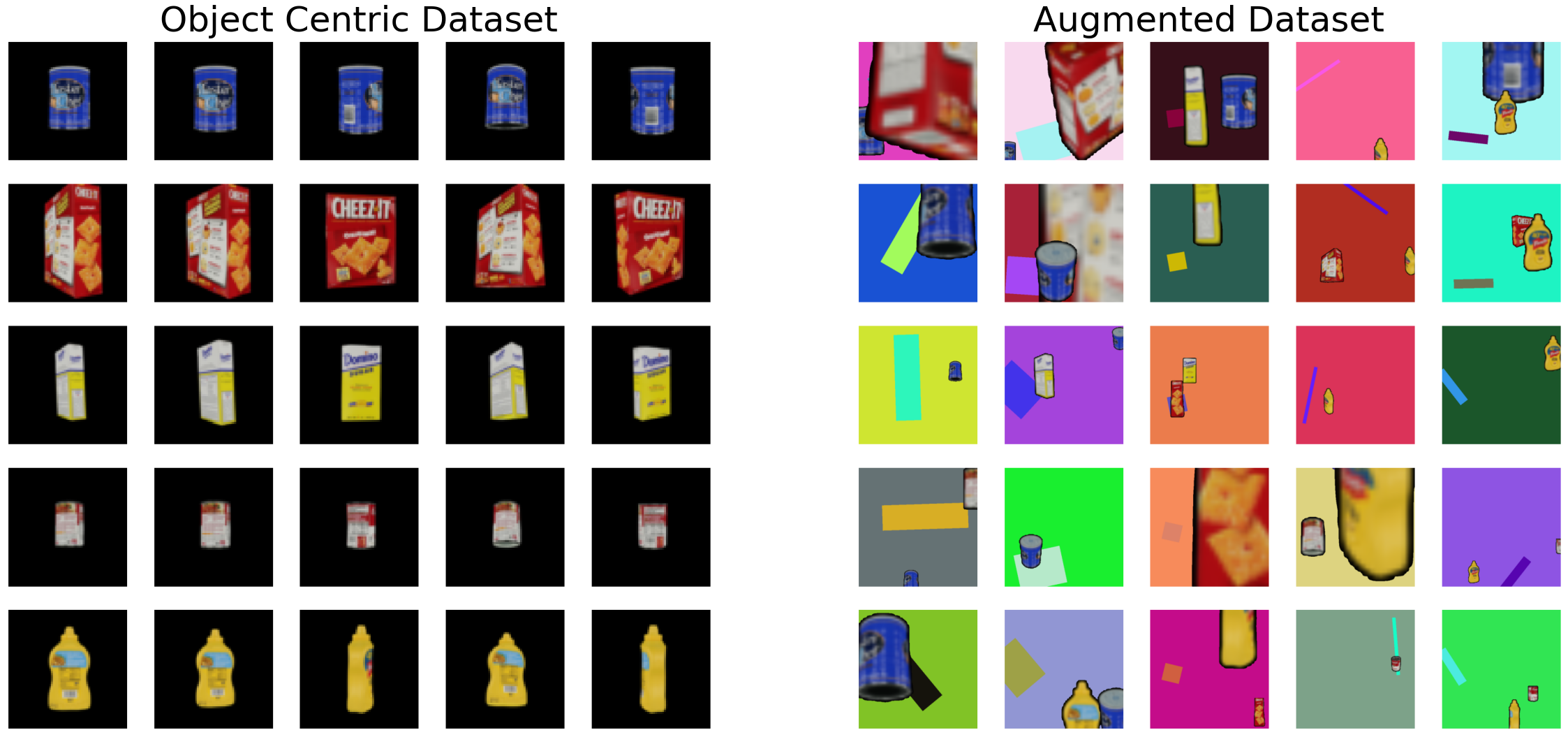}
    \caption{A few examples from the object-centric dataset (left) and the augmented dataset (right).}
    \label{fig:dataset}
\end{figure}

We only consider observing objects from a fixed distance, as this will allow an agent to use the scale of an object for estimating the distance to the object. We then sample all observations from a fixed distance of 40 cm from the object center. Points are sampled uniformly on a sphere around around the object center, considering a fixed radius of 40 cm. Both the RGB observation and the viewpoints are recorded. For each object category, a  dataset of 500 samples is created. Some samples from this dataset are shown on the left of Figure~\ref{fig:dataset}.

For training the second phase that considers scene (not object-centric) observations, we do not create a novel dataset. Instead, we augment and combine our object-centric observations to lie in a similar distribution to full scenes. To this end, we randomly scale and translate the object in the observation to simulate a sense of translation. Then two randomly oriented rectangles in different colors are used as a background, which increases robustness to different environments. On top of this, we also add a different random object from one of the other categories that is also scaled and translated according to the same distribution as the main object. This sometimes occludes the object of interest and makes our model more robust. Finally, to add augmentation for training the object identity prediction, in 20\% of the cases we remove the object from the observation. Some samples from this dataset are shown in Figure~\ref{fig:dataset}.

\section{Neural Network Parameters}
\label{appendix:nn}

There are multiple neural networks working together for the approximate inference of the hidden variables. When the agent observes the scene, first a fully convolutional neural network estimates which pixels belong to the object of category k, this is the observation mask $\mathbf{\alpha}_{k,t}$. The parameterization of this layer is shown in Table~\ref{tab:mask}.

\begin{table}[h]
    \centering
    \caption{configuration of the Masking Neural Network.}
    \begin{tabular}{lc}
        \toprule[1.5pt]
        \textbf{Layer} & \textbf{Output Neurons/Filters} \\
        \midrule[1.5pt]
         Interpolate to 64x64 & - \\
         Convolutional (1x1) + Leaky ReLU & 16 \\
         Convolutional (3x3) + Leaky ReLU & 32 \\
         Convolutional (3x3) + Leaky ReLU & 64 \\
         Convolutional (3x3) & 1 \\
         Median Pool (5x5) & - \\
         Sigmoid & -\\
         Interpolate to 480x480 & - \\
         \bottomrule[1.5pt]
    \end{tabular}
    \label{tab:mask}
\end{table}

The observation is then masked using a pixel-wise product $\mathbf{o}_{t} \odot \mathbf{\alpha}_{k,t}$ and further processed to predict object identity $\mathbf{i}_{k}$, i.e. whether the object of interest is present in the observation. This neural network predicts the parameters for a spatial transformer, i.e. the center of the object in pixel space and the object scale directly. Its parameterization is shown in Table~\ref{tab:stn}. This directly predicts the scale and normalized center of the object of interest. To get the predicted scale in a range for the model to optimize we scale and invert this output $h$ of the neural network using the following transform: $\sigma_{k,t} = 1 / (3.8 \cdot h + 0.2)$. 

\begin{table}[h]
    \centering
    \caption{Configuration of the Spatial Transformer Network.}
    \label{tab:stn}
    \begin{tabular}{lc}
    \toprule[1.5pt]
    \textbf{Layer} & \textbf{Output Neurons/Filters} \\
    \midrule[1.5pt]
    Interpolate to 32x32 & - \\
    Flatten & - \\ 
    Linear + Leaky ReLU  & 256 \\ 
    Linear + Leaky ReLU & 128 \\
    Linear + Leaky ReLU & 64 \\
    Linear + ELU & 3 \\
    \bottomrule[1.5pt]
    \end{tabular} 
\end{table}

Using the output of the spatial transformer network, the masked observation can be processed into an object-centric observation after which this is processed by two neural networks. One predicts the object identity $\mathbf{i}_k$ or whether the object of interest is present in the observation. The parameterization is shown in Table~\ref{tab:identity}. 

\begin{table}[h]
    \centering
    \caption{Configuration of the Identity Estimation Network.}
    \begin{tabular}{lc}
        \toprule[1.5pt]
        \textbf{Layer} & \textbf{Output Neurons/Filters} \\
        \midrule[1.5pt]
        Convolutional (4x4) + Leaky ReLU & 8 \\
        Convolutional (4x4) + Leaky ReLU & 16 \\
        Convolutional (4x4) + Leaky ReLU & 32 \\
        Convolutional (4x4) + Leaky ReLU & 64 \\
        Flatten & - \\ 
        Linear + Sigmoid & 1 \\ 
         \bottomrule[1.5pt]
    \end{tabular}
    \label{tab:identity}
\end{table}

Finally, a CCN is able to estimate a latent variable describing the object pose. This model consists of three neural networks. First, an encoder takes an object-centric observation as an input and predicts the parameters of a multivariate Gaussian with 8 dimensions and with a diagonal covariance matrix. This variable represents the object pose in an object-centric reference frame. The details of this model can be found in Table~\ref{tab:ccnencoder}. The final output of this model is a vector of twice the latent size, representing the mean and standard deviation of the multivariate Gaussian. A decoder or likelihood model takes as input a sample from this pose distribution and decodes it into a pixel-based observation. Using the transition model, the distribution over the pose can be estimated for future viewpoints. The model takes as input the concatenated vector of a sample from the current belief over pose and the action. The action in this case is a two-dimensional vector representing the displacement in azimuth and elevation, as we only consider observations from a fixed distance. 

\begin{table}[h]
    \centering
    \caption{Configuration of the CCN Encoder.}
    \begin{tabular}{lc}
        \toprule[1.5pt]
         \textbf{Layer} & \textbf{Output Neurons/Filters} \\
        \midrule[1.5pt]
         Convolution (4x4) + Leaky ReLU & 8 \\
         Convolution (4x4) + Leaky ReLU  & 16 \\ 
         Convolution (4x4) + Leaky ReLU & 32 \\
         Convolution (4x4) + Leaky ReLU & 64 \\
         Flatten & - \\
         Linear & $2 \cdot 8$ \\
         Softplus for $\sigma$ output & -  \\
         \bottomrule[1.5pt]
    \end{tabular}
    \label{tab:ccnencoder}
\end{table}

\begin{table}[h]
    \centering
    \caption{Configuration of the CCN Decoder.}
    \begin{tabular}{lc}
        \toprule[1.5pt]
        \textbf{Layer} & \textbf{Output Neurons/Filters} \\
        \midrule[1.5pt]
        Linear & 4096 \\
        Unflatten to 64x8x8 & - \\
        Convolution (5x5) + Leaky ReLU & 64 \\
        Interpolate to 17x17 & - \\ 
        Convolution (5x5) + Leaky ReLU & 64 \\
        Interpolate to 35x35 & - \\ 
        Convolution (6x6) + Leaky ReLU & 32 \\
        Interpolate to 69x69 & - \\ 
        Convolution (6x6) + Leaky ReLU & 16 \\
        Convolution (1x1) & 3 \\
        \bottomrule[1.5pt]
    \end{tabular}
    \label{tab:ccndecoder}
\end{table}

\begin{table}[h]
    \centering
    \caption{Configuration of the CCN Transition Model.}
    \begin{tabular}{lc}
        \toprule[1.5pt]
        \textbf{Layer} & \textbf{Output Neurons/Filters} \\
        \midrule[1.5pt]
        Linear + Leaky ReLU & 128 \\
        Linear + Leaky ReLU & 256 \\
        Linear + Leaky ReLU & 256 \\
        Linear & $2 \cdot 8$ \\
         Softplus for $\sigma$ output & -  \\
         \bottomrule[1.5pt]
    \end{tabular}
    \label{tab:ccntransition}
\end{table}


These neural networks are trained in two distinct steps, using the same object-centric dataset of 500 observation-viewpoint pairs per object. Drawing inspiration from object-centric learning in infants~\cite{smith_active_2021}, the first phase focuses on learning object-centric representations. While the second phase focuses on learning to decompose the scene into object-centric representations. In the first phase, the CCN models is optimized on the $\mathcal{L}_1$-loss, described in Equation~\ref{eq:l1}. This is done using the Adam~\cite{kingma_adam_2017} optimizer with the configuration parameters shown in Table~\ref{tab:ccnoptim}. In the second phase, the other models are optimized by minimizing the 
$\mathcal{L}_2$-loss (Equation~\ref{eq:l2}), again using the Adam~\cite{kingma_adam_2017} optimizer, but now with a learning rate of $1\cdot10^{-4}$. 

\begin{table}[h]
    \centering
    \caption{Hyperparameters for optimization of the CCN in the first stage of training.}   
    \label{tab:ccnoptim}
    \begin{tabular}{lcc}
        \toprule[1.5pt]
        & \textbf{Initial Value} & \textbf{Range} \\
        \midrule[1.5pt]
        $\lambda_1$ & 40 & $[80, 100]$  \\
        $\lambda_2$ & 40 & $[80, 100]$  \\
        $\lambda_3$ & 40 & $[80, 100]$  \\
        $\lambda_4$ & 10 & fixed  \\
        Adjust frequency & 500 steps & - \\
        Adjust factor & 1.01 & - \\
        Learning rate & $5\cdot10^{-4}$ & - \\
        \bottomrule[1.5pt]
    \end{tabular}
\end{table}

\section{Details on Active Agents} 

When we consider the agent driven through expected free energy. A new target viewpoint $\mathbf{v}_{t+1}$ is computed every 10 steps, or unless the target is reached. 

When estimating the 3D multivariate Gaussian, the mean is placed at the estimated depth along the negative z-direction of the camera pose in OpenGL format. The depth is acquired using the following relation with the scale: $d = 0.4 / \mathbf{\sigma}_{k,t}$, as observations are trained for a distance $d=0.4$. The variance is then set at a value of $0.1973/2$ along the depth dimension, and $0.02$ along the other dimensions.

The particle filters for estimating the position of the individual objects are initialized with 10k particles. At each step, the particles are resampled with a standard deviation of $0.025 cm$. When the agent does not observe the object in question, all particles that are in view of the camera and closer than the nearest (estimated) object, are set to a low value of $10^{-5}$ before normalization. 

For the implementation, the reader is referred to the supplementary materials.

\clearpage

\section{Additional Results}
\label{appendix:results}

\begin{figure}[h]
    \centering
    \includegraphics[width=0.95\textwidth]{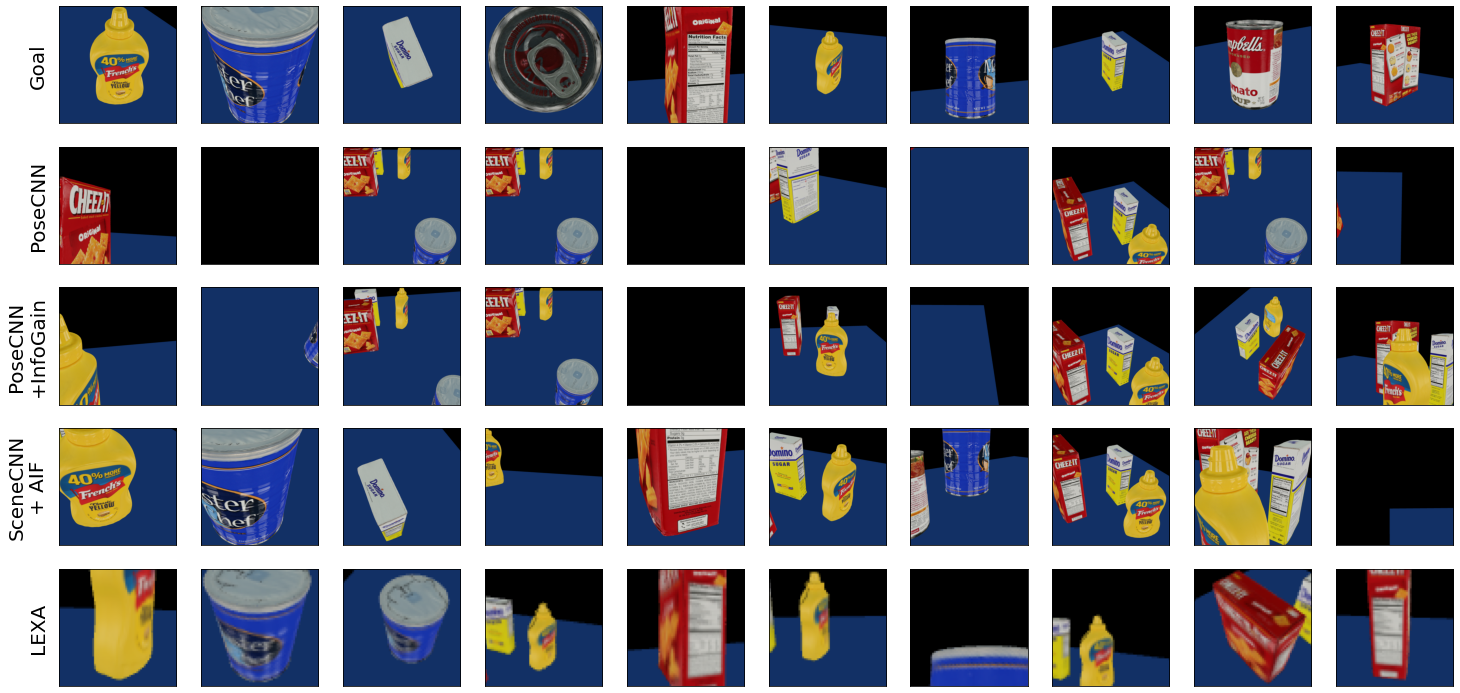}
    \caption{Qualitative results of the scene in which LEXA was trained. In this configuration, five objects are randomly placed on a table, and the agent must reach a goal observation. The top row shows the goal observation, while the other rows show the final reached observation for each agent. The simulation is stopped after 350 steps.}
    \label{fig:additionalresults}
\end{figure}

All approaches are evaluated on a set of ten goals from the environment in which the LEXA agent was trained. The qualitative results can be observed in Figure~\ref{fig:additionalresults}. The goal observation is displayed in the top row, while the other goals show the final reached frame of the agent with a maximum of 350 steps. 

\end{document}